\documentclass[]{interact}
\usepackage{hyperref}
\hypersetup{
  pdfproducer={},
  pdfcreator={},
}
\usepackage{amsmath,amssymb,amsfonts}
\usepackage{algorithmic}
\usepackage{graphicx}
\usepackage{textcomp}
\usepackage{subcaption}
\usepackage{caption}
\usepackage{breqn}
\usepackage{amsmath}
\usepackage{array,ragged2e}
\usepackage{stackengine}
\usepackage{multirow}
\usepackage{xurl}
\usepackage[backend=biber,style=ieee-alphabetic,sorting=nyvt]{biblatex}
\begin{document}

%\articletype{ARTICLE TEMPLATE}% Specify the article type or omit as appropriate

\title{High Performance P300 Spellers Using GPT2 Word Prediction With Cross-Subject Training}

\author{
\name{
Nithin Parthasarathy\textsuperscript{a}, James Soetedjo\textsuperscript{b}, Saarang Panchavati\textsuperscript{c}, Nitya Parthasarathy\textsuperscript{d}, Corey Arnold\textsuperscript{c}, Nader Pouratian\textsuperscript{e} and William Speier\textsuperscript{c}
}
\affil{\textsuperscript{a}Dept of Comp. Science,University of Illinois at Urbana Champaign;
\textsuperscript{b}Department of Bioengineering, University of Washington;
\textsuperscript{c}Department of Radiological Sciences, UCLA;
\textsuperscript{d}PE Investments, Boston, MA, (formerly at Department of Computer Science, MIT);
\textsuperscript{e}Department of Neurological Surgery, University of Texas, Southwestern, Dallas
}
}
\maketitle

\begin{abstract}
Amyotrophic lateral sclerosis (ALS) severely impairs patients' ability to communicate, often leading to a decline in their quality of life within a few years of diagnosis. The P300 speller brain-computer interface (BCI) offers an alternative communication method by interpreting a subject's EEG response to characters presented on a grid interface.

This paper addresses the common speed limitations encountered in training efficient P300-based multi-subject classifiers by introducing innovative "across-subject" classifiers. We leverage a combination of the second-generation Generative Pre-Trained Transformer (GPT2) and Dijkstra's algorithm to optimize stimuli and suggest word completion choices based on typing history. Additionally, we employ a multi-layered smoothing technique to accommodate out-of-vocabulary (OOV) words.

Through extensive simulations involving random sampling of EEG data from subjects, we demonstrate significant speed enhancements in typing passages containing rare and OOV words. These optimizations result in approximately $ 10\% $ improvement in character-level typing speed and up to $ 40\% $ improvement in multi-word prediction. We demonstrate that augmenting standard row/column highlighting techniques with layered word prediction yields close-to-optimal performance.

Furthermore, we explore both "within-subject" and "across-subject" training techniques, showing that speed improvements are consistent across both approaches.

\end{abstract}


\section{Introduction}
Amyotrophic lateral sclerosis (ALS) is a progressive neurodegenerative disease involving motor neurons in the cerebral cortex, severely impairing patients' lives \cite{ALS}. ALS patients currently have a means to communicate through non-invasive brain-computer interfaces (BCI) \cite{gao}-\cite{mcfarland2} which allow direct translation of electric, magnetic, or metabolic brain signals into control commands of external devices. The P300 speller is a common BCI communication interface that presents stimuli to produce an evoked response. In this technique, a character matrix is presented before the subject and their EEG response (more specifically, the P300 evoked response potential or ERP) which varies based on the highlighted display characters is recorded, processed, and interpreted.\cite{mcfarland2}  

The classical paradigm for P300-based BCI speller was introduced by Farwell and Donchin in 1988 \cite{farell}. The Row-Column (RC) paradigm is the most popular speller format, which consists of a $ 6\times 6 $ matrix of characters. We will refer to this matrix or GUI as a {\it “flashboard”} in this paper. This flashboard is presented on the computer screen and the row and columns are typically flashed/highlighted in a random order, with the subject selecting a character by focusing on it. The flashing row or column evokes a P300 ERP response in the EEG and therefore, the classifier algorithm can determine the target row and column after averaging several responses, thereby selecting the desired character. Although this work also uses a $ 6 \times 6 $ row/column flashboard design, the results readily extend to any other type of flashboard, such as the checkerboard \cite{townsend} format.

\begin{figure*}[!t]
\centering
\includegraphics[width=5.6in,height=3.2in]{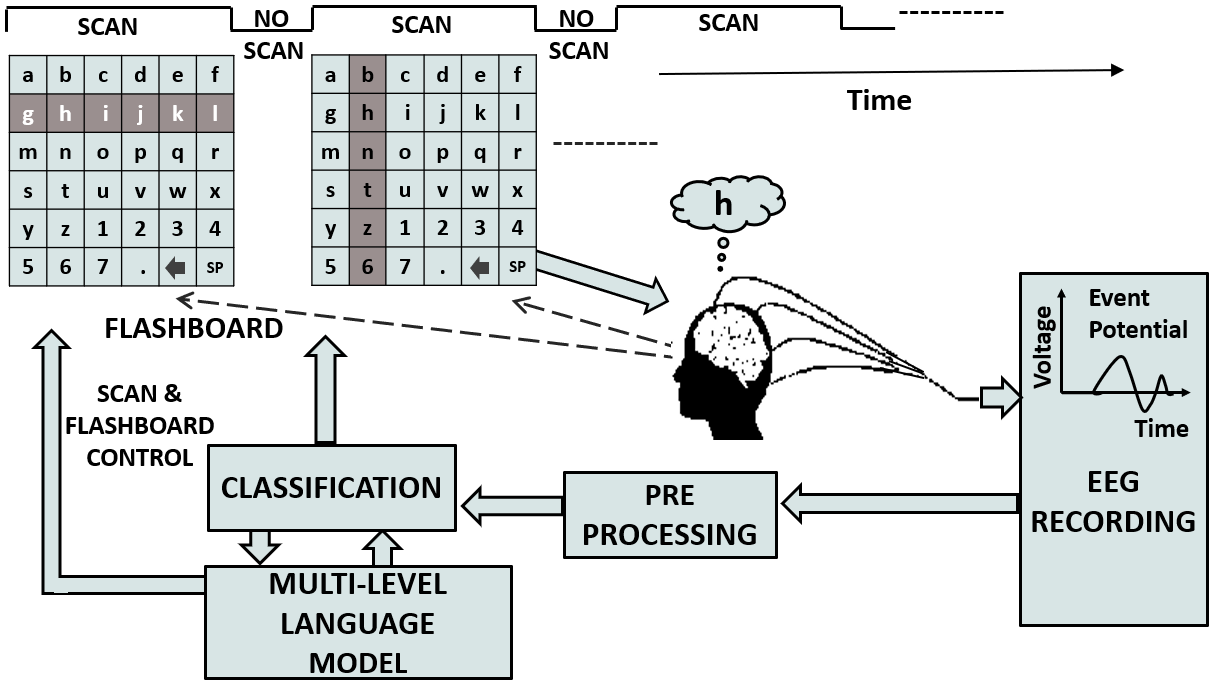}
\caption{Block diagram of the P300 speller with the subject EEG response to character highlights on a flashboard.}
\label{blockdiag}
\end{figure*}

Typing speed remains a significant limitation of the P300 speller (a block diagram of which is shown in Figure \ref{blockdiag}), along with the challenge of efficient classification across multiple subjects. This paper addresses both issues uniquely and comprehensively, leveraging advanced language models for word prediction while also introducing novel {\it “across”} subject classifiers. These innovations which invoke new cross-disciplinary approaches simplify EEG response decoding and enhance typing speed significantly. Furthermore, this paper also demonstrates a novel methodology to seamlessly upgrade existing spellers with these new techniques without hardware changes. 

% Note that most of the new techniques demonstrated in this paper are generic enough to apply to other BCI-evoked response methods too.

While the P300 speller is designed to be a communication medium, signal classification methods traditionally have not taken complete advantage of existing knowledge about the language domain and its associated correlations. As a simple example of such dependencies, note that it is more likely that the character {\it 'e'} is followed by {\it 's'} than {\it 'z'} in an English word \cite{mayzner}. Historically, while natural language has been utilized for years to improve classification in other areas such as speech recognition \cite{jelinek}, its use in the BCI field \cite{speier1},\cite{speier2},\cite{nithin1} is a relatively recent movement. While language models provide a means of exploiting the natural redundancies in speech \cite{speier_spell}, character selections in prior work have largely been treated as independent elements chosen from a set without prior information. More specifically, current techniques of random highlighting ignore multi-word context \cite{kindermans1},\cite{kindermans2} along with word associations and word prediction given prior and partially formed words, thereby limiting communication efficiency. In fact, BCI studies using n-gram language models provide a poor representation of natural language as by disregarding context, they can potentially attach a high probability to character strings that do not formulate words. Therefore, in contrast to prior work, this research exploits the additional contextual information available through increasingly sophisticated \cite{Kucera},\cite{openai} language models to increase BCI system efficiency.

The goal of this study is multifold, with a central theme of eliminating redundancy with sophisticated contextual word prediction algorithms,  allowing for efficient with and {\it “across”} subject classifier training. Such classifiers minimize subject-dependent calibration requirements, thereby facilitating the development of a single universal classification scheme. 
%Maybe
We used extensive offline simulations to robustly evaluate this approach and its potential in online studies. Using simulations based on multi-subject EEG data along with layered word completion algorithms such as GPT2 in conjunction with Dijkstra's algorithm, P300 speller performance is dramatically improved in both {\it “within”} and {\it “across”} subject classifiers. New flashboard and scanning configurations are also developed to further optimize the speller performance and evaluate their relative merits. To overcome some limitations in prior work, smoothing algorithms are employed to enhance the out-of-vocabulary (OOV) performance of word as well as character prediction.

The simulation techniques developed in this work allow comparison of many speller scenarios over large, and representative targets. These include OOV words that would otherwise be infeasible in an online study because of the significant time requirements. 
% The organization of this paper is as follows. In section \ref{section_methods}, detailed description is provided on the various techniques and algorithms used for the flashboard, stimuli, and classifiers as well as character and word prediction. Section \ref{section_results} illustrates the results, while Section \ref{section_discussion} provides a summary of the results, which is followed by the conclusions and future directions section.
\section{Methods}
\label{section_methods}
\subsection{Language models}
In spoken English, it is well established that vowels occur more frequently than consonants \cite{cornell}. Hence, from a timeline perspective of when they get highlighted in the flashboard scanning order, it is only natural that they occur earlier than some relatively rare consonants such as $ 'x' $ or $ 'z' $. We will now describe how to perform such trade-offs using multi-level language models. 

\begin{figure*}[!t]
\centering
\includegraphics[width=5.6in,height=2.6in]{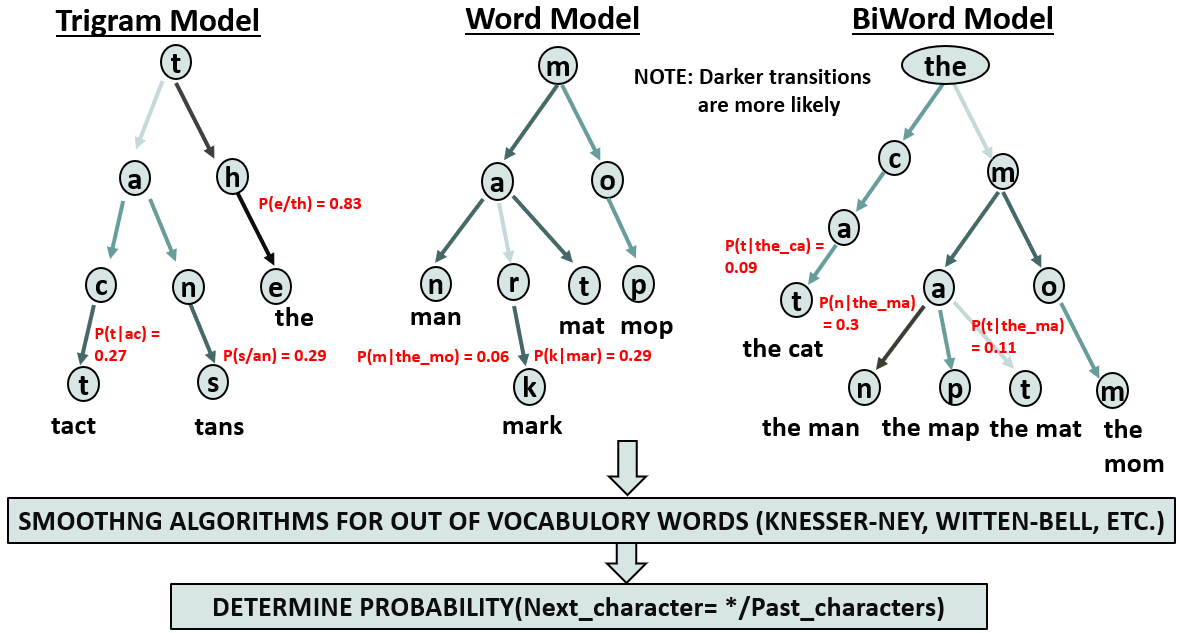}
\caption{Types of language models used. Starting with the simplest on the left to the most complex on the right.}
\label{fig_langmodel}
\end{figure*}

Smoothing techniques \cite{jurafsky}, allow transitioning across
models when OOV words are encountered. Consequentially, complex models can be blended in with simpler models (Figure \ref{fig_langmodel}). As an example, consider the word {\it ’cat’} and suppose {\it ’ca’} has already been spelled. Assume that {\it ’cat’ } is not present in the language model. 
Hence, the probability of {\it 't'} given that {\it 'ca'} has already been spelled (represented by the notation, {\it p(‘t’|’ca’)} is zero. Smoothing updates this OOV probability by considering other contexts where {\it 'a'} and {\it ‘t’} have co-occurred through {\it p('t'/'a’)}. 
Multiple smoothing techniques were considered, and Knesser-Ney smoothing was chosen for its superior OOV \cite{kinderman3} performance. Also note that in general, the more sophisticated the language model, the better the conditional probability can be predicted \cite{openai}. Language models such as the one used contain a large vocabulary, which results in high prediction accuracy even as character length increases.

\textbf{Knesser-Ney smoothing:} In the following terminology, {\it Model('*')} denotes the occurrence frequency of the characters {\it'*'}. For example, {\it biword\_model('the\,quick') } is the frequency of the word {\it 'the\,quick'} in the biword language model. Also, as an example of conditional probability syntax, {\it p\_biword ('c' | 'the\,qui')}  is the biword model probability of the character {\it 'c'} being next given that the partial phrase {\it 'the\,qui'} has been decoded.
\begin{multline}
p\_biword ('c' | 'the\,qui') = \max (biword\_model ('the\,quic')\\ - d_1, 0) / biword\_model ('the\,qui') \\
+  d_1 \cdot L_1 \cdot p\_word('c' | 'qui') 
\label{kn_eqn1}
\end{multline}
where $ L_1 $is the normalizing constant equal to the number of distinct letters that follow {\it 'the\,qui'} in the biword model divided by the biword model count of {\it'the\,qui'}. If {\it biword\_model ('the\,qui' ) = 0, $ L_{1}$ =1} to take care of out of vocabulary (OOV) conditions.
\begin{multline}
p\_word ('c' | 'qui') = \max\left(word\_model ('quic') - d_2, 0\right) \\
/ word\_model(\text{'qui'}) + d_2 \cdot L_2 \cdot p\_trigram ('c' | 'qui') 
\label{kn_eqn2}
\end{multline}
\begin{multline}
p\_trigram('c' | 'ui') = \max\left(trigram\_model ('uic') - d_3, 0\right) \\
/ trigram\_model(\text{'ui'}) + d_3 \cdot L_3 \cdot p\_bigram('c' | 'i')
\label{kn_eqn3}
\end{multline}
\begin{multline}
p\_bigram ('c' | 'i') = \max\left(bigram\_model ('ic') - d_4, 0\right) \\
/ bigram\_model(\text{'i'}) + d_4 \cdot L_4 \cdot p\_unigram('c')
\label{kn_eqn4}
\end{multline}
%\begin{multline}
\begin{equation}
p\_unigram ('c') = unigram\_model ('c') / unigram\_model ('') 
\label{kn_eqn5}
%\end{multline}
\end{equation}
where {\it unigram\_model('')} = total model entries and $d_1, d_2, d_3, d_4$ are tunable parameters with $ 0 \leq d_1, d_2, d_3, d_4 \leq  1$. All these {\it d} values were set to the commonly used mid-range default of 0.5. Just as $ L_1$ is a biword model normalizing constant in Equation \ref{kn_eqn1} whose derivation was described earlier with an example, $ L_2, L_3, L_4$ are also normalizing constants similarly obtained from the corresponding lower order language models. More specifically, in the example used, $L_2$ is the normalizing constant equal to the number of distinct letters that follow {\it 'qui'} in the word model divided by the word model count of {\it'qui'}. Further, for satisfying OOV conditions, if {\it word\_model ('qui') = 0, $L_2$ =1}. If {\it trigram\_model ('ui') = 0, $L_3$= 1} and if {\it bigram\_model ('i') = 0, $L_4$ = 1} thereby accounting for the values of $L_1$...$L_4$ in all OOV cases.  
In summary, through Equations \ref{kn_eqn1}-\ref{kn_eqn5}, the most complex model in Figure \ref{fig_langmodel} falls back to lower-level models with OOV characters.

\subsection{Feature Extraction: Training Methodology}
\label{subsection_feature_extraction}
The step-wise linear discriminant analysis (SWLDA) as described in Speier \cite{speier2} et al. and originally used in \cite{farell} was employed as the classifier. SWLDA utilizes a discriminant function, which is determined at the training step \cite{draper}. The SWLDA classifier was trained using one of two methods:  Across subject cross validation (ASCV) (also known as leave-one-subject-out  cross  validation) and within-subject cross validation (WSCV). Though both use SWLDA, note that these two techniques fundamentally differ in how the entire subject responses are split between what was used for training and was used for testing. \\
\textbf{a) ASCV:} Out of $ N $ subjects, one subject was chosen to be the test subject, while the other $ N-{\it 1 }$ subjects’ data were used to train the SWLDA classifier. The SWLDA classifier then attempted to intake the test subject’s data and spell out the target phrase. Note that this technique finally produces one classifier table for all $ N $ subjects.\\
\textbf{b) WSCV:} In the WSCV, 2-fold cross-validation using SWLDA was performed on each individual subject EEG data (unlike ASCV) resulting in a unique classifier output for each subject. More specifically, the output phrase contained at least $ 20 $ characters and had 120 flashes for every character that was split into three sets. During the first testing iteration, the first set was chosen to be the test phrase, while the other two sets were used to train the SWLDA classifier. On completion, the same procedure would repeat, but with the second and third set being the test phrase during the second and third trial respectively. This method would then be replicated on all the subjects.

During training, class labels were predicted using ordinary-least-squares. Next, the forward step-wise analysis adds the most important features, and the backward analysis step determines and removes the least important features. This process repeats until the number of features remains constant after a set number of iterations or until the number of required features is met. The final selected features are then added to the discriminant function. The score for flash $ i $ for character $ t $, $ y_t^i $ , can then be computed as the dot product of the feature weight vector with the features from that trial’s signal.

\subsection{Virtual flashboard design using a language model}
The flashboard is a central part of a P300-based BCI system. In this section, we will describe schemes that organize the highlighted flashboard characters based on their occurrence probabilities. Note that these schemes are mapped {\it virtually} onto conventional static flashboards, whereby only the highlighted set of characters are modified and not the underlying flashboard itself. In other words, the flashboard is constant through the flashing sequence, though the highlighted characters can be arbitrarily chosen. By doing so, this readily overlays onto any flashboard in current use

\begin{figure*}[!t]
\centering
\includegraphics[width=4.8in,height=1.6in]{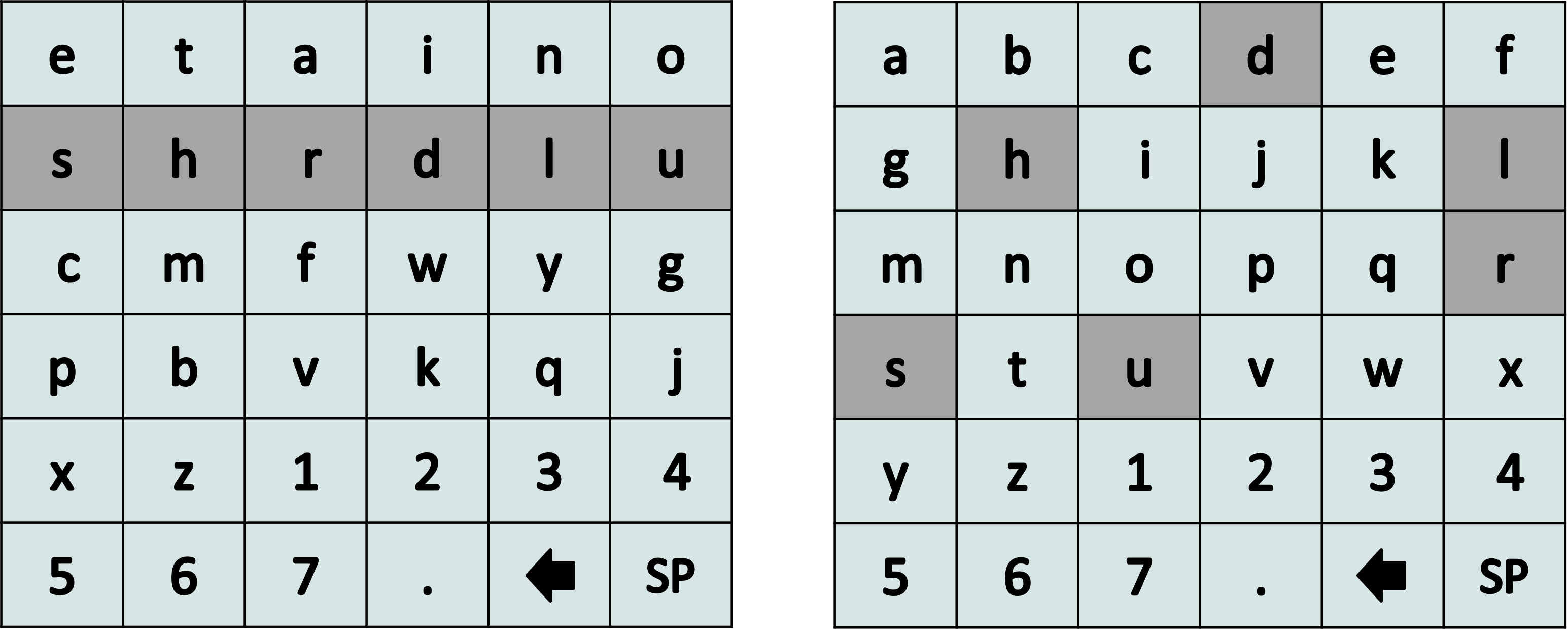}
%\caption*{({\it a})     \hspace{7cm}   ({\it b})}
\caption{a) Probabilistic flashboard highlighting order in a sequential frequency-sorted flashboard. {\it SP} denotes word space. b) Sequential flashboard characters shown in Figure \ref{fig_seq}a are {\it virtually} mapped onto a conventional alphabetical flashboard}
\label{fig_seq}
\end{figure*}

\begin{figure*}[!t]
\centering
\includegraphics[width=4.8in,height=1.6in]{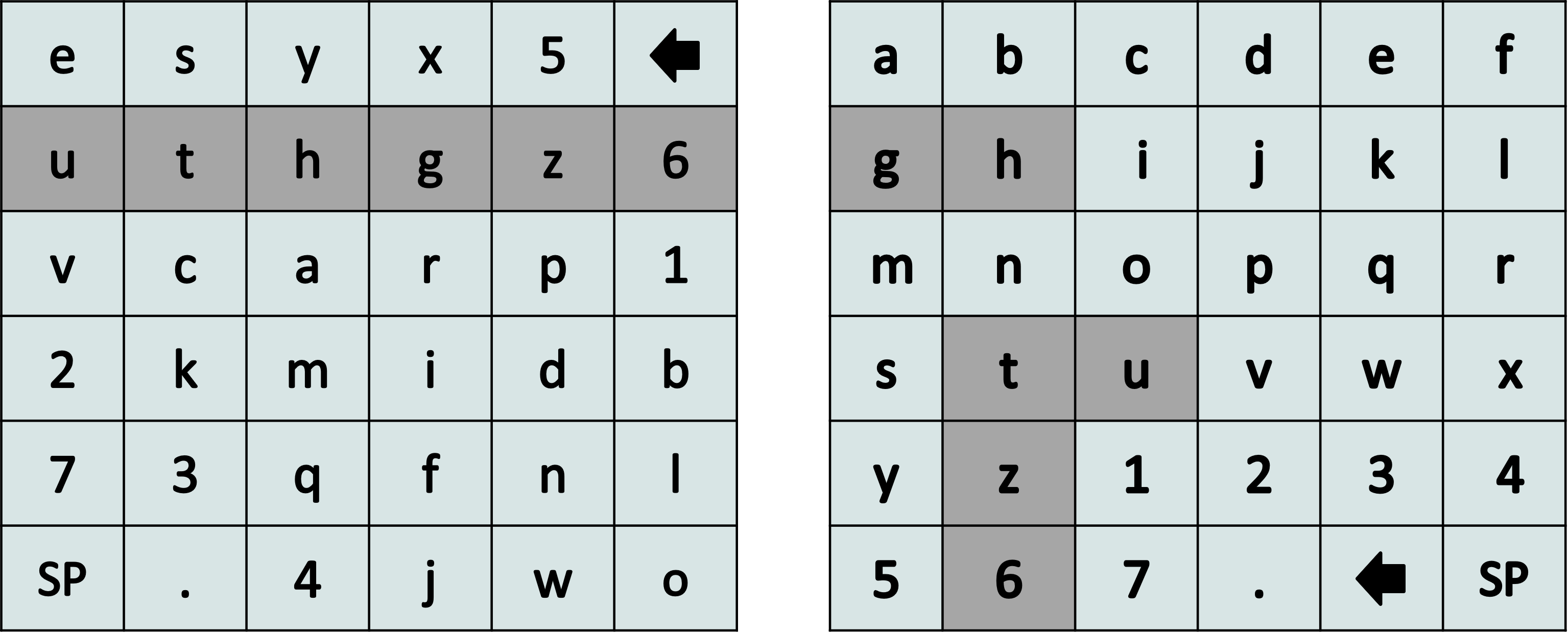}
%\caption*{({\it a})     \hspace{7cm}   ({\it b})}
\caption{(a) Diagonal flashboard highlighting where most likely characters {\it e, t, a, i ...} are organized along diagonals.
b) Diagonal highlighted flashboard characters in Figure \ref{fig_diag}a are {\it virtually} mapped onto a conventional alphabetical flashboard}
\label{fig_diag}
\end{figure*}

\begin{figure}
\centering
%\begin{subfigure}[b]{0.3\textwidth}
\begin{subfigure}[b]{0.4\columnwidth}
\includegraphics[width=1.6in,height=1.8in]{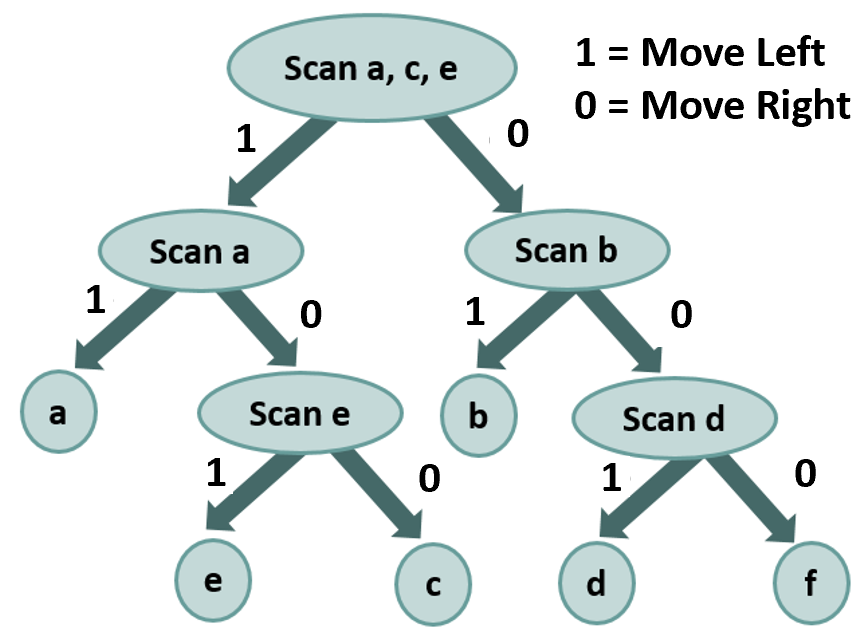}
\caption{}
\end{subfigure}%
%\begin{subfigure}
\begin{subfigure}[b]{0.4\columnwidth}
\includegraphics[width=1.6in,height=1.6in]{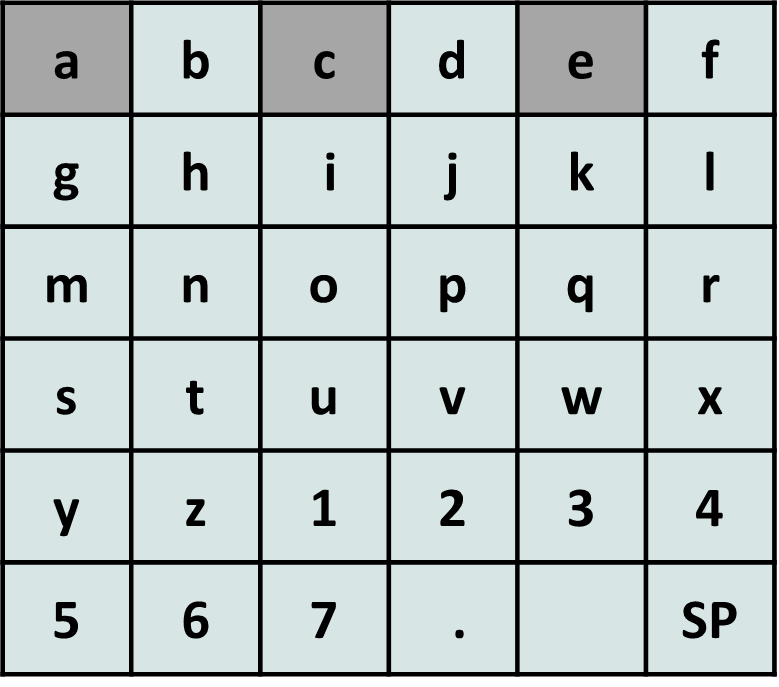}
\caption{}
\end{subfigure}
\caption{a) Huffman scanning steps where {\it 1} and {\it 0} are the direction to move b) {\it Virtual} mapping onto a conventional alphabetical flashboard.}
\label{fig_huffman}
\end{figure}

% , including popularly used random highlighting. In related work, Townsend \cite{townsend} propose the checkerboard paradigm where flashing occurs in non-adjacent groups as opposed to row-column to reduce errors. \\
\textbf{a) Sequential flashboard:} The flashboard highlighting is {\it virtually}  modified (as shown in Figure \ref{fig_seq}) such that each set of highlighted characters are picked in a frequency weighted rather than alphabetical manner. More specifically, higher frequency characters are highlighted earlier, whereby the initial flashes are more likely to select the target character. Results from these modifications will be described in section \ref{section_results} and discussed in section \ref{section_discussion}. \\
\textbf{b) Diagonal flashboard:} Consider the diagonal design of Figure \ref{fig_diag} where the most likely highlighted characters are organized along diagonals. In the figure, note that the most likely letters {\it e, t, a, i, n, o} are on the main diagonal followed by the next likely letters on upper and lower diagonals and so on. This design forces characters that appear in similar contexts to flash separately, thus making them easier to distinguish. Once again, this {\it virtual} technique can also easily overlay on to static alphabetically arranged flashboards. Gains from these schemes are provided in section \ref{section_results}. \\
\textbf{c) Huffman flashboard:} 
Huffman codes \cite{huffman} are popular in lossless data compression and used in prominent multimedia standards such as JPEG and MP3. The fundamental idea is to use fewer bits to represent the most frequent characters and more bits to represent less frequent characters. By doing so, the overall number of bits required to represent the complete data is minimized, approaching information theoretical performance bounds. Note that different frequency tables would result in different Huffman codes. This section deals with adapting this approach to P300 spellers. For example, suppose we had only $ 5 $ symbols. Huffman scanning descends through a probabilistically organized binary tree (as illustrated in Figure \ref{fig_huffman}) based on the prior decoded character.

One requirement of Huffman flashing is that the determination of whether the subject responded positively to the highlighted flashboard characters must be feedback to the flashing unit to help it identify the next set of flashing characters. For feasibility, the overall decoder latency should be small, which creates a limitation for its practical use. 

\subsection{Further optimization of the scanning order}
This section describes modifications to adapt the P300 BCI scanning to optimize performance.\\
\textbf{a) Weighted scanning order:} Conventional random and deterministic flashing are straightforward schemes, where the flashing set of characters is either picked deterministically or round-robin. In contrast, in the proposed weighted flashing enhancement to conventional flashing, the highlighted set is picked based on that set's probability as opposed to the random scheme where the sets are assumed to be equi-probable. The idea is that the more likely characters are highlighted earlier.\\
\textbf{b) Dynamic stopping:} Once a character is determined at the decoder, continuing the flashing creates an unnecessary reduction in speed, which is avoided by using a {\it“dynamic stopping”} strategy. Flashing of the current character is terminated and the next character flash is begun.
\begin{figure*}
\centering
\includegraphics[width=5.6in,height=2.5in]{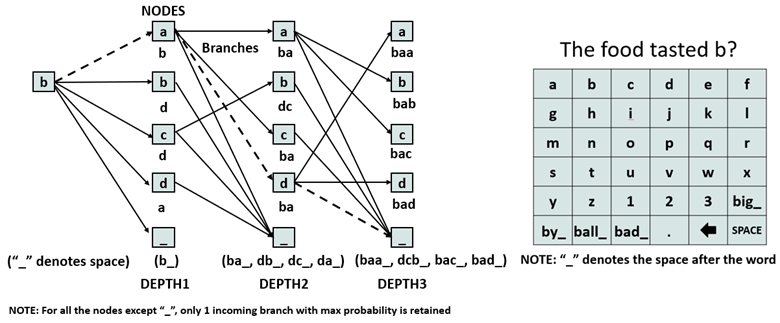}
\caption*{\hspace{2cm}({\it a})     \hspace{8cm}   ({\it b})}
\caption{(a) Implementation of Dijkstra's algorithm to find most likely word completions. (b) Flashboard shows most likely word completions where 4 characters are replaced by the word choices. Note: “\textunderscore” represents space between words.}
\label{dijkstra_diag}
\end{figure*} 
\subsection{Word prediction algorithms and methodology}
Flashboard efficiency can be increased by replacing unlikely characters with word choices. \\
\textbf{a) Dijkstra’s Algorithm:} We used Dijkstra’s algorithm \cite{Dijkstra} for word prediction. This was implemented in the form of
\begin{figure}
\centering
\includegraphics[width=2.4in,height=1.8in]{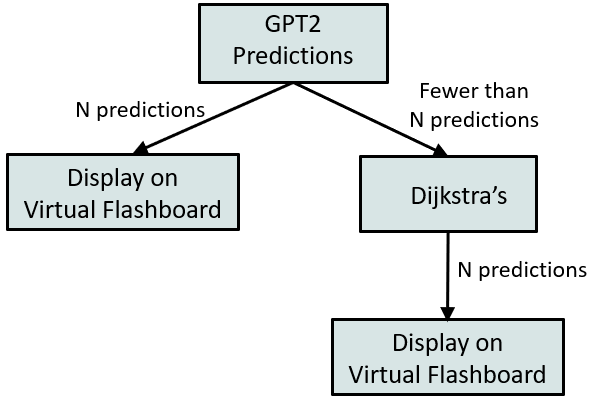}
\caption{Layered GPT2 word choices to obtain {\it N } most likely word completions with fallback to Dijkstra's algorithm}
\label{fig_layered_gpt2}
\end{figure}
% The classic technique for traversing a tree or a graph is a depth-first search or a breadth-first search, which are well known algorithms \cite{efreq}. Though fast, it does not provide a technique for evaluating OOV words which occur when the partially spelled word is not present in the language model. In contrast, 
 a trellis similar to dynamic programming techniques as shown in Figure \ref{dijkstra_diag}. As shown in the figure, every node of a trellis stage has $ M $ incoming possibilities (where $ M $ is the number of unique characters in the flashboard) out of which only the one with the highest probability is retained. In doing so, the complexity was reduced, as every possibility did not have to be exhaustively evaluated. Further, as more stages were incorporated into the trellis, longer-length word choices are obtained while resulting in higher evaluation time and complexity. For each branch at a given stage, probabilities were determined by using the frequency count of branch characters given prior characters using the language model along with smoothing. At every stage, the highest probability completions then provided the most likely word completions.\\
\textbf{b) GPT2:} A large unsupervised transformer \cite{openai},\cite{transformer} based model released by {\it OpenAI}, GPT2 is an enhanced deep neural network follow-on to the earlier GPT model and is powerful enough to generate coherent passages of text. In contrast to earlier models, GPT2 uses “attention mechanisms” to focus on segments of input text it thinks are the most relevant. In the context  of BCI, it can become a sophisticated tool that predicts the subsequent word in a partially formed sentence given prior words. Though GPT2 is powerful, it does not have the ability of OOV prediction. However, it can be combined with Dijkstra’s algorithm (as shown in Figure \ref{fig_layered_gpt2}) to account for OOV words, thereby creating a layered approach similar to character-based smoothing algorithms. More specifically, in case GPT2 word choices result in an empty or partially complete set of choices, one can then move to Dijkstra’s to predict OOV words.

\subsection{Decoder}
A simple decoding scheme based on thresholding is used in this work to deduce the character or word. First, the score for each flash is computed by, $ y_t^i  = {\mathbf w}{\bf .z_t^i }$ where the feature weight vector is determined by feature extraction as described in the previous subsection. With the assumption that the distributions were Gaussian \cite{speier2}, the {\it “attended”} and {\it “non-attended”} signals (note that this terminology refers to highlighted and non-highlighted characters on the flashboard) were found and given by 
\begin{dmath}
 f(y_{t}^{i}/x_{t}) = \left\{
    \begin{array}{ll}
         \frac{1}{\sqrt{2\pi\sigma_{a}^{2}}} 
   e^{\frac{(y_{t}^{i} - \mu_{a})^2}{2\sigma_{a}^2}} & \mbox{if } x_{t} \in {\mathbf A_{t}^i}\\
   \frac{1}{\sqrt{2\pi\sigma_{n}^{2}}} 
   e^{\frac{(y_{t}^{i} - \mu_{n})^2}{2\sigma_{n}^2}} & \mbox{if } x_{t} \notin {\mathbf A_{t}^i}\\     
    \end{array}
\right.
\label{eqnPdf}
\end{dmath}    
where $ {\mathbf A_{t}^i} $ is the set of characters illuminated for the $ i^{th} $ flash for character $ t $ in the sequence. $ \mu_a,\sigma_a^2,\mu_n,\sigma_n^2 $ are the means and variances of the distributions for the {\it attended} and {\it non-attended} flashes, respectively. 
\begin{dmath}
 P (x_t|{\bf y_t},x_{t-1},...,x_0  )= 
 \frac{P (x_t | x_{t-1},...,x_0)} {P ({\bf y_t} | x_{t-1},..,x_0  )).P ({\bf y_t} |x_{t},..,x_0  )}
 \end{dmath}
 \begin{dmath}
 =\frac{1}{Z}.P (x_t | x_{t-1},..,x_0  )\prod_if(y_t^i | x_t)
\end{dmath}
where $ P(x_t|x_{t-1},.., x_0) $ is the prior probability of a character given the history, $ f(y_t^i|x_t) $ are the pdf’s from Equation \ref{eqnPdf}, and $ Z $ is a normalizing constant. The language model is used to derive the prior as per
\begin{dmath}
P(x_t | x_{t-1},..,x_0  )=   Count(x_t,x_{t-1},..,x_0  ) )/\sum_{x_t}Count(x_{t},x_{t-1},..,x_{0})
\end{dmath}
where $ Count(x_{t},x_{t-1},..,x_0) $ is 
the number of occurrences of \begin{math} x_{t}x_{t-1}..x_{0} \end{math} in the corpus.
A threshold probability, $ P_{Thresh} $, is then set to determine when a decision should be made. The program flashes characters until either $ max_{x_t} \,P (x_t|{\bf y_t}, x_{t-1},.., x_0) \geq P_{Thresh} $ or the number of sets of flashes reaches a predetermined maximum value. The classifier then selects the character that satisfies $ arg\, max_{x_t}\, P (x_t|{\bf y_t}, x_{t-1},.., x_0) $.

The subject can correct typing errors by invoking the backspace, which is provided in the flashboard. As a result, all the typing errors do get fixed, making the overall error rate zero. Alternatively, one can have an autocorrect wherein typed words are checked for presence in a dictionary and erred words are replaced by closest most likely words. This technique is faster, but some mistyped words are likely to either be mis-corrected or left uncorrected, which would result in a non-zero final error rate. While a vast array of more powerful decoding schemes involving hidden Markov models, machine learning, and particle filters \cite{speier2} can be used, note that these enhancements only complement the techniques introduced in this paper and do not deter the results. 

\subsection{Simulation overview}
Data for offline analyses were obtained from 78 healthy volunteers with normal or corrected to normal vision between the ages of 20 and 35. All data were acquired using GTEC amplifiers, active EEG electrodes, and electrode cap (Guger Technologies, Graz, Austria); sampled at {\it 256\,Hz}, referenced to the left ear; grounded to {\it AF_Z}; and filtered using a band-pass of {\it 0.1 – 60\,Hz}. The electrode set consisted of 32 channels placed according to a previously published configuration \cite{jessicalu}. The system used a {\it 6×6} character grid, row and column flashes, and a stimulus duration of 100\,msec and an interstimulus interval of 25\,ms for a stimulus onset asynchrony of 125\,msec.  
Statistical variation from such small data sets would limit the confidence in the results. As a result, a large and meaningful dataset, the text of the {\it “Declaration of Independence”} (DOI) is used as the simulation data \cite{doi}. The full length of this document is 7,892 characters after removing all punctuations and special characters. The DOI was chosen as the simulation dataset as it is concise, well-known, and contains words of varying length, including a number of out-of-vocabulary (OOV) words that stress the simulation. Furthermore, to keep the simulation accurate, numerous subject responses are used to provide a wide range of brain activity. 

% With this, we believe that our results establish a fundamental basis for future work, which could further supplement and enhance these results with additional variation in subject material as well as brain activity.
 
While running the simulations with the DOI dataset, the {\it “attended”} and {\it “unattended”} scores (the terminology refers to whether the desired character is present or absent in the highlighted flashboard character) for a given subject were chosen from the lab results of that subject. The lab results are sampled by a two state Markov chain which models the {\it “attended”} and the {\it “unattended”} states. Thus, there are four possible sampling combinations depending on whether the current and previous state was {\it “attended”} or {\it “unattended”}. When a character or word is decoded, it is compared to the intended character or word. In the case of an error, a {\it “Backspace”} is initiated as the subsequent character. Following this, the deleted character/word is flashed again. The net effect of this is to {\it “Undo”} errors, thereby making the final error rate zero. However, note that this could potentially even require multiple attempts and a long follow-up time. One could significantly constrain the follow-up time and tolerate the residual error, and this evaluation could be a possible future study.
\subsection{Performance evaluation metrics}
\label{itr_section}
Information Transfer Rate (ITR) is a crucial metric in BCI evaluation, quantifying the speed and accuracy of communication \cite{mcfarland1}. The metric is appealing for several reasons: it is derived from information theory principles, it combines the competing statistics of speed and accuracy, and it reduces the performance to an information transfer problem that can be compared across applications \cite{pierce}.
It is typically calculated using the formula: $ \frac{log_{2}(N+1)}{T+P_{f}(T_r+T_c)} \times 60 \; bits/minute $ where $ N $ is the number of characters in the speller (grid size), $ T $ is the total time taken to complete the task (including pauses, corrections, etc.). Note that $ P_f $ is the probability of incorrectly selecting a character, $ T_r $ is the time required for a single character selection, and $ T_c $ is the time taken for error corrections.

A second metric we used is the retry rate (or error rate) which measures the number of backspaces initiated as a fraction of the overall communicated characters. Note that when a character or word is decoded in error, the next character is assumed to be a backspace. The objective being to attempt mimicking a subject, correcting the misinterpreted error or word. This process repeats until the character/word is decoded correctly   resulting in the error rate for that character or word going to 0. However, when the simulation of a particular character or word requires more than 75 flashboard scans, it is abandoned and an ITR of 0 (100\% error rate) assigned for that particular character or word and the simulation moves on to the next character or word. 

Some errors (e.g., obvious typos) do not affect the readability of text and could therefore be allowed to remain in the final output without affecting the meaning. Allowing these errors to remain could then potentially result in a more efficient system because it would not waste time forcing the user to make these corrections. Language models could also be used to automatically correct some of these errors  \cite{speier2},\cite{speier3} if they are not corrected manually. 

Both the Shapiro-Wilk test \cite{nonparametric} and the Kolmogorov-Smirnov test \cite{nonparametric} on the results comprehensively rejects normality $( p < 10^{-3} )$. Consequentially, use non-parametric tests for hypothesis testing. The Kruskal-Wallis test \cite{nonparametric} was performed on the data, $( p < 10^{-6} )$ rejecting the null hypothesis of each scheme's ITR, error rate originating from the same distribution. Hence, the Wilcoxon signed rank test \cite{nonparametric} for paired samples is then conducted to determine the statistical significance.

\section{Results}
\label{section_results}
\begin{figure*}
\begin{subfigure}[b]{0.5\columnwidth}
\centering
\includegraphics[width=2.8in,height=3.0in]{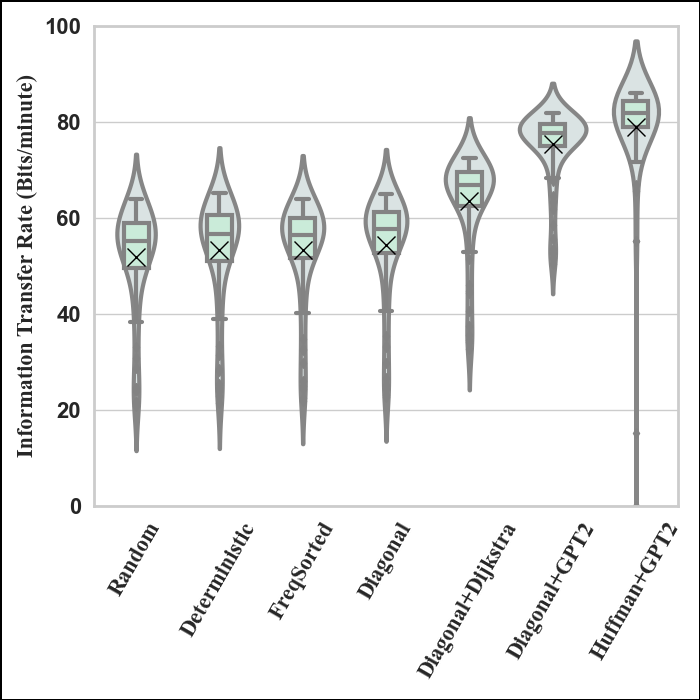}
\caption{Within Subjects}
\end{subfigure}%
\begin{subfigure}[b]{0.5\columnwidth}
\centering
\includegraphics[width=2.8in,height=3.0in]{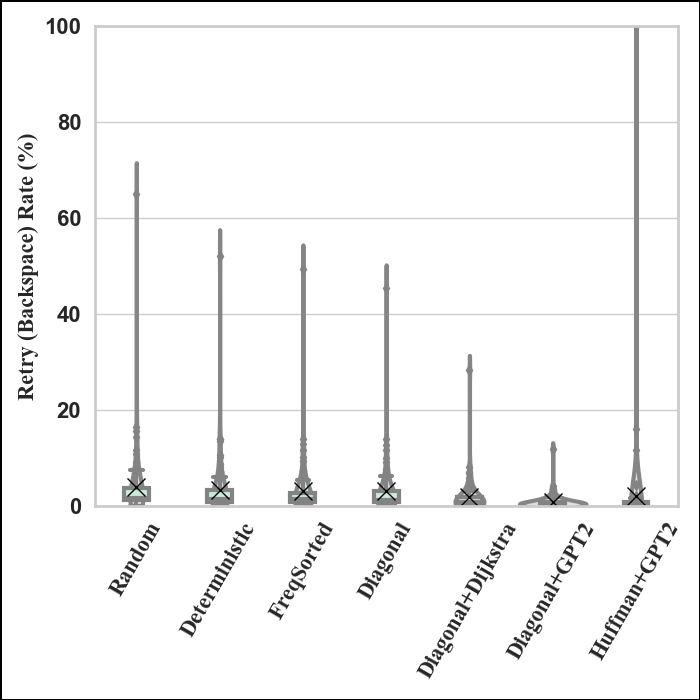}
\caption{Across Subjects}
\end{subfigure}
\caption{a) ITR performance comparison of different schemes for the {\it “Within Subjects”} (WSCV) training b) Error rate \\
\textbf{NOTE:} Violin plot containing an embedded box-and-whiskers that shows the minimum, first quartile, median, third quartile, and maximum data. {\it ‘X’} is the mean ITR.}
\label{wscv_only}
\end{figure*}
Figures \ref{wscv_only}a and \ref{wscv_only}b show the performance of different schemes for 78 subjects along with their error rates in a {\it violin} plot along with an embedded traditional box and whiskers plot. The violin plot depicts the density of the data in the form of a histogram, while the embedded box and whiskers plot provides relevant statistics. 
\begin{figure}
\begin{subfigure}[b]{0.49\columnwidth}
\centering
\includegraphics[width=2.77in,height=3.0in]{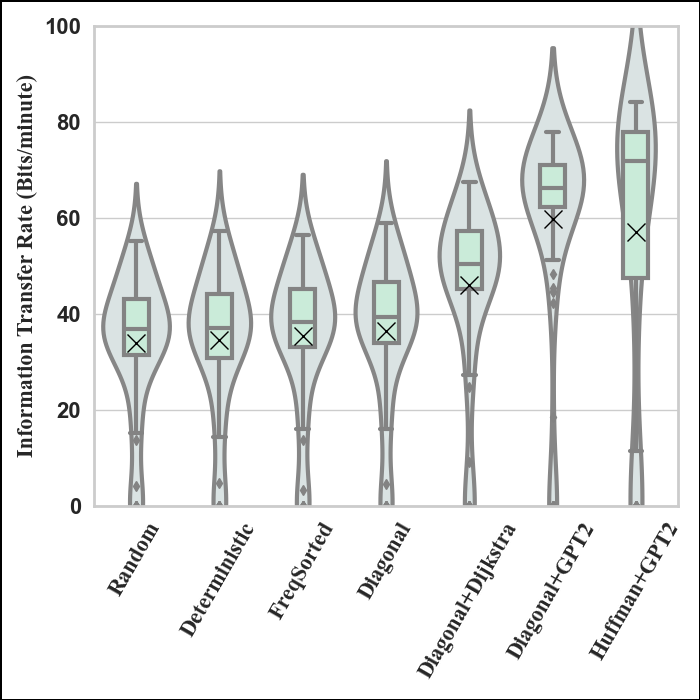}
\caption{}
\end{subfigure}
\begin{subfigure}[b]{0.5\columnwidth}
\centering
\includegraphics[width=2.8in,height=3.0in]{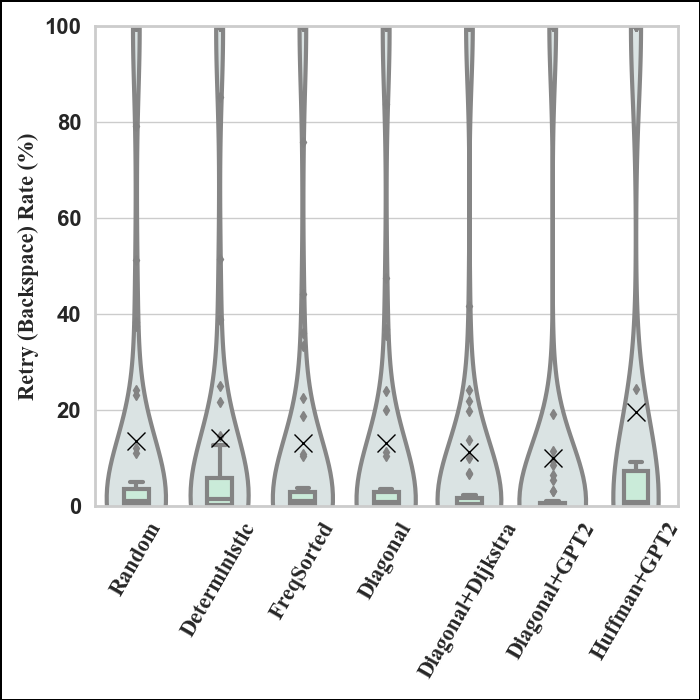}
\caption{}
\end{subfigure}
\caption{a) ITR performance comparison of different schemes for the {\it “Across Subjects”} (ASCV) training b) Error rate\\
}
\label{ascv_only}
\end{figure}
\begin{table}
\large
\centering
\begin{tabular}{|c | c  | c|} 
 \hline
 \multicolumn{1}{|c|}{\multirow{2}{*}{\textbf{ Scheme}}} & 
 \multicolumn{1}{c|}{\textbf{ WSCV ITR}} &
  \multicolumn{1}{c|}{\textbf{ ASCV ITR}}\\
   & (bits/min)	& (bits/min) \\ 
 \hline 
 \hline
Random	          & $ 51.93 \pm 11.2 $	&   $ 33.95  \pm 14.4  $  \\ [0.25ex]\hline
Deterministic     &	$ 53.42 \pm 11.4 $          &	$ 34.49  \pm 14.9       $\\ [0.25ex]\hline
FreqSorted        &	$ 53.44 \pm 10.8 $          &	$ 35.44  \pm 15.0     $\\ [0.25ex]\hline
Diagonal          &	$ 54.43 \pm 11.0 $          &	$ 36.51  \pm 15.4     $\\ [0.25ex]\hline
Diagonal+Dijkstra &	$ 63.59 \pm 10.1 $          &	$ 46.05  \pm 17.8    $\\ [0.25ex]\hline
Diagonal+GPT2     &	$ 75.31 \pm 7.5 $          &	$ {\bf 59.75 \pm 21.0}  $\\ [0.25ex]\hline
Huffman+GPT2      &	$ {\bf 78.86 \pm 13.0} $     &	$ 57.06  \pm   30.1   $\\ [0.25ex]\hline
\end{tabular}
\caption{ITR (mean $\pm$ standard deviation format) for WSCV and ASCV subject training. Scheme with the highest mean ITR is shown in bold.}
\label{table1}
\end{table}
In this case, the classifier was trained on each subject individually (WSCV training of subsection \ref{subsection_feature_extraction}). A maximum of six word suggestions were provided in this simulation, with numbers in the flashboard replaced with the word suggestions. In the Figure \ref{wscv_only}b, the error rate captures the backspace rate due to a wrong character decoding. However, note that the final error rate in the output text is zero in all these cases. Further, ITR was calculated as defined in section \ref{itr_section}. There is a wide range of performance based on the scheme employed, with the  minimum average ITR for WSCV of about 51.9 bits/minute, while the maximum is 75.3 bits/minutes with GPT2. Equivalent numbers for the ASCV scheme range from 34.5 to 59.75 bits/minute.

 The Huffman+GPT2 configuration provided the highest average WSCV ITR (78.86 bits/minute), which was significantly higher than all other configurations other than Diagonal+GPT2 (75.31 bits/minute, p=0.01). While the Huffman scheme provided the highest gains, it also had the highest standard deviation of 13 bits/minute, which is higher than the Diagonal+GT2 which had a standard deviation of 7.5 bits/minute. The width of the violin plots also illustrates the standard deviation in the results through the width of the histogram for each scheme. In the ASCV scenario, the same trend holds in with the Diagonal+GPT2 configuration provided the highest ITR (59.75 bits/minute) as seen in Figure \ref{ascv_only} and Table \ref{table1}. Note that this figure shows the same results as Figure \ref{wscv_only}, but for the ASCV training of subsection \ref{subsection_feature_extraction}. Once again, the Huffman+GPT2 combination has the highest variance of 30.1 bits/minute, which is much higher than all the other schemes.
 
\begin{figure*}
\begin{subfigure}[b]{0.5\columnwidth}
\centering
\includegraphics[width=2.8in,height=3.0in]{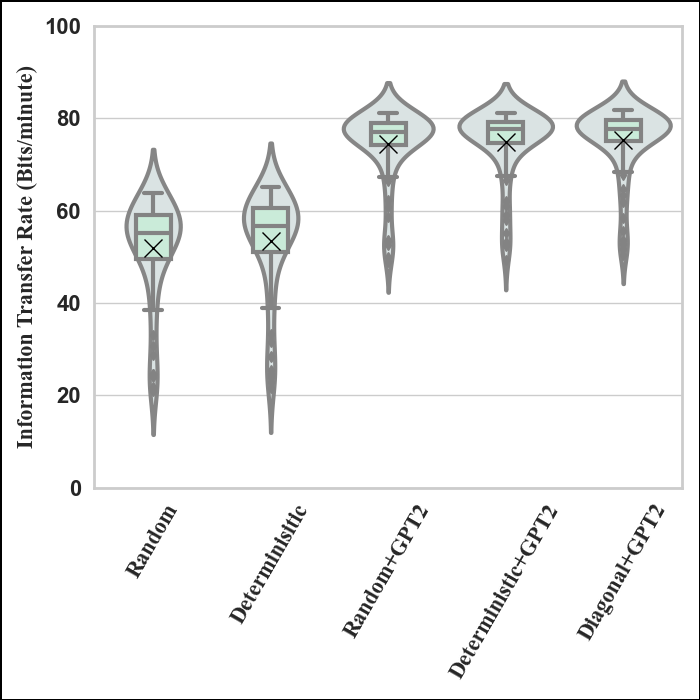}
\caption{Within Subjects}
\end{subfigure}%
\begin{subfigure}[b]{0.5\columnwidth}
\centering
\includegraphics[width=2.8in,height=3.0in]{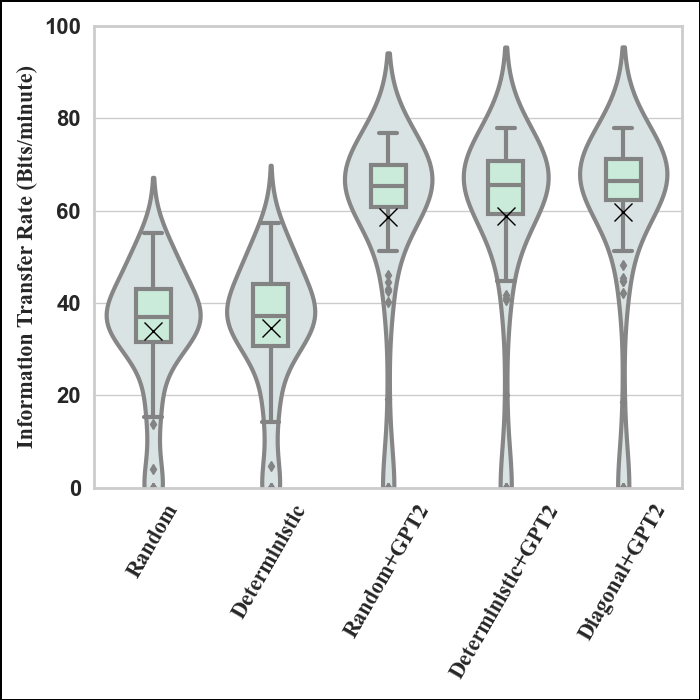}
\caption{Across Subjects}
\end{subfigure}
\caption{Violin plot of the information transfer rate of conventional flashboards, showing the effect of word completion on performance improvement for {\it a) “Within Subjects”} (WSCV) and {\it b) “Across Subjects”} (ASCV). \\ \textbf{NOTE:}{\it “Regular”} refers to the traditional random scan with alphabetical flashboard while {\it “Regular-Wcomp”} refers to the same but with word suggestions included}
\label{wscv_ascv}
\end{figure*}

Figure \ref{wscv_ascv} shows the performance of traditional schemes when they are augmented with word completion and prediction. As shown in Figure \ref{fig_seq}, using the concept of virtual flashboard, word completions schemes become a simple overlay onto conventional flashboards. Hence, one then observe the performance gains achieved by using this word completion retrofit technique. As an example, using WSCV, the mean ITR increases from about 53.4 to 75.3 bits/minute when GPT2 is using in conjunction with “Random” or “Deterministic” schemes, a gain of about 41\%. With ASCV, the mean ITR increases from 34.5 to 59.75 bits/minute providing even larger gains of about 73\%.
\section{Discussion}
\label{section_discussion}
In this study, a large set of subject EEG responses (which represent a wide range of evoked responses) was used to simulate and optimize some novel flashboard highlighting strategies. Note that these simulations were run on a standard HP laptop using an Intel i7 processor with 16 GB memory. 
In the within-subject (WSCV) analysis, the frequency-sorted and diagonal schemes outperform the traditional ones. For example, the mean ITR for the diagonal scheme was $ 54.4 $ bits/minute, which showed a relatively modest improvement   over the random scheme ($ 51.9 $ bits/minute, $ p < 0.001 $). The diagonal scheme is superior to the traditional schemes in all metrics including the minimum and mean ITR as well as lower error rate and better error statistics. Again, by the virtual flashboard technique shown in Figures \ref{fig_seq},\ref{fig_diag}, the underlying flashboard itself is unchanged. Rather, it is the highlighting of the characters which varies with the particular scheme. Hence, this should have a negligible impact on the subject's attention. 

For the same example in the previous paragraph, when Dijkstra’s algorithm alone is used for word completion, the ITR gain increases to $ 63.6 $ bits/minute. When GPT2 is used along with Dijkstra's, the same gain further improves to $ 75.3 $ bits/minute.
The combination of Huffman and GPT2 increased the average ITR to 79.0 bits/minute, which is significantly higher than all other configurations ($ p < 0.001 $, figure \ref{wscv_only}a). However, several subjects had poor ITR values using this method due to high error rates (Figure \ref{wscv_only}b). This result is not surprising, given that when there are errors, the Huffman scheme frequently traverses down the incorrect branch. Then, based on how far the Huffman sequence has progressed, it could potentially require either a small or large number of flashes to begin all over and repeat the process. Consequently, with a large error rate, this effect creates a large variance in the average number of flashes required per character. The diagonal+GPT2 configuration had the second highest ITR (75.0 bits/minute) which was significantly higher than all methods other than Huffman+GPT2 ($ p < 0.001 $). Once gain, note that the retry rate is also much lower with this method, as seen in Figure \ref{wscv_only}b. On average, both of these methods provided nearly $ 20\% $ better ITR than any of the other schemes. The closest performance is with Diagonal+Dijkstra. In general, using word completion algorithms provide substantial gains over schemes that use character prediction only, which is consistent with earlier work \cite{speier_spell}. Further, by using smoothing, flashboard character probabilities for 17 biwords in the DOI dataset which were OOV in the biword language model were obtained by transitioning to the word model and trigram models. Note that improvement is proportional to the complexity of the language model, with GPT2 standing out in performance. 

Across subjects (ASCV) performance (Figure \ref{ascv_only}a) shows a similar trend as WSCV. However, in comparison to WSCV, there are more subjects who have a lower ITR and higher error rate (Figure \ref{ascv_only}b). The diagonal character flashboard showed an $ 7.6\% $ improvement over the regular flashboard with random flashing, while the Dijkstra’s and GPT2 word completion provide $ 35.6\% $ and a massive $ 76\% $ gain respectively. In this case, the GPT2 also outperforms the Huffman scheme. Also, there was a vast improvement in the retry rate when word completion is employed. Interestingly, the gains from word prediction are not language model dependent, though the choice of the specific model does influence the relative gains. Further, feed-forward schemes with word completion provide a good performance trade-off (in comparison to feedback schemes such as Huffman) with relatively lower variation in performance as seen from the distribution spread in the Violin plots.

Retrofitting, which refers to augmenting standard schemes with word prediction, increases mean ITR to within $ 2\% $ of the best-in-class performance (diagonal+GPT2) for both within and across subjects (Figure \ref{wscv_ascv}). This result was extremely insightful as it suggested that one could preserve the traditional approaches, upgrading them with a word completion technique based on an advanced language model along with smoothing and obtain close to optimal performance. It also indicates the tremendous gain when powerful language models are employed, while showing that word completion using the model provides far superior gain as compared to that with just character-level flashboard improvements.

The main limitation in this study pertains to the assumptions in the simulation. As with all offline studies, it assumes that neural signals are not affected by the presence of feedback. While this is a strong assumption, it is supported by previous studies that have shown consistent performance in online and offline analyses \cite{speier2}. However, it is possible that some changes presented here could have an additional effect on a subject's neural signals, which will need to be evaluated in online studies. For example, the Huffman approach changes the flashing frequency (as compared to other methods) which has been shown to potentially reduce P300 amplitude. In addition, the simulation assumes that the order of the stimuli does not matter and that the signals will not change over time. These two assumptions may potentially be weakened by factors such as varying attention and user fatigue, resulting in gradual changes in the EEG response over time. 

% We agree that online experiments are important, and have indeed included them in many of our prior papers \cite{speier2},\cite{speier3}. However, there are several factors involved in the current work which make online testing impractical in this study. The goal of the model proposed in this study was to make the system more flexible, allowing for rare and out of vocabulary words while testing of system components that could previously only be demonstrated online. Further, it narrows the scope of tedious BCI experimental evaluation through optimization of parameters over long simulation data sets. In our earlier online studies, subjects generally chose short sentences comprised of common English words, which would not demonstrate the value of the proposed models. We could instead choose a passage for our subjects, but if we imposed a sentence with rare and/or OOV words, it would likely bias analysis in favor of our model. Hence, we decided that the best way to evaluate the performance of our proposed model would be to choose a long passage that is separate from our training corpus so that it is likely to contain rare or OOV words at a rate more in line with general English text. 

In this work, we primarily focused on offline studies to highlight the performance gain provided by our model. A big part of the novelty of this study is the ability to more robustly type OOV words. In previous offline studies \cite{speier2}, \cite{speier3}, subjects generally chose much shorter sentences comprised of common English words, which would not sufficiently demonstrate the efficacy of our approach. Those studies also found that the improvements shown in simulations were recapitulated in an online setting. While it would have been possible to choose a longer passage for our subjects with several OOV words, it would likely bias analysis in favor of the model. we decided that the best way to evaluate the performance of our proposed model would be to choose a long passage that is separate from our training corpus so that it is likely to contain rare or OOV words at a rate more in line with general English text. Though simulation is an imperfect representation of BCI performance, it proved to be a useful tool to assess the efficacy of our approach. In future studies, we will conduct an online study to assess the real world impact of the approach. This research employed the GPT-2 large language model (LLM). Since the conclusion of this work, superior models like GPT-3 and GPT-4 have been released. We expect that we can further enhance these results through the use of increasingly advanced LLMs, providing more context-dependent word suggestions. 
% Since the Declaration of Independence was too long to realistically be typed during a BCI session, we were required to design the simulation environment to represent how the system could work for a large population of BCI subjects. While we acknowledge that simulation is an imperfect representation of BCI performance, we will point out that in all of our previous studies, offline performance gains from NLP and ML improvements have held up in online sessions \cite{speier2},\cite{speier3}. Although online experiments are definitely important, we believe that this offline simulation is sufficient for the current study, and online analysis can be reserved for a future longitudinal study that can provide a larger body of testing output text.
% While we believe that the GPT-2 is a powerful language prediction tool, there are certainly contextual limitations when dealing with language generation \cite{openai}. However, advances in the GPT based language models such as GPT-3 \cite{gpt3} mitigate such limitations, further enhancing its value. Hence, we 
% do not see this as a fundamental limitation to the results of this paper. In fact, it opens up opportunities to experiment with new models which are more context-sensitive.
\section{Conclusion and future directions}
Language models used for improving classification speed and accuracy in the P300 speller can also be effectively utilized to create whole-word suggestions for predictive spelling. New techniques for flashboard construction based on probabilistic modeling were introduced in this work. In combination with word completion algorithms, typing speed is significantly improved over current random highlighting techniques. Using real-world data from $ 78 $ subjects, simulations with a large dataset were shown to increase typing speed (over standard schemes) in within-subject training from $ 51.9 $ bits/minute to $ 75.4 $ bits/minute. An across-subject classifier methodology was also introduced, which eliminated the need for subject-specific calibration. The same ITR metric improved from $ 33.9 $ bits/minute to $ 59.7 $ bits/minute in across subject training with word prediction. In addition, the error or retry rate is also reduced, closing the gap between theory and practice. Further improvements can potentially be realized by using dynamic thresholding techniques where threshold settings can be adapted based on subject response and will be the study of future work along with online experiments.

\end{document}